\documentclass[conference]{IEEEtran}

\usepackage{cite}          % numbered, compressed citations, e.g. [1], [2]--[5]
\usepackage{graphicx}
\usepackage{amsmath,amssymb}
\usepackage{multirow}
\usepackage{booktabs}
\usepackage{array}
\usepackage[table]{xcolor}
\usepackage{url}
\usepackage{hyperref}

\newcommand{\citep}[1]{\cite{#1}}
\newcommand{\citet}[1]{\cite{#1}}

\newcommand{\best}[1]{\textbf{#1}}
\newcommand{\second}[1]{\underline{#1}}

\graphicspath{{figures/}}

\title{JADE-GS: Joint Allocation of Deblurring Evidence for Event-Assisted 3D Gaussian Splatting}

\author{%
\IEEEauthorblockN{Haoyu Fu\IEEEauthorrefmark{1}\IEEEauthorrefmark{2}, Jiafeng Huang\IEEEauthorrefmark{1}\IEEEauthorrefmark{2}, Yuchen Wang\IEEEauthorrefmark{1}\IEEEauthorrefmark{2}, Shengjie Zhao\IEEEauthorrefmark{3}}
\IEEEauthorblockA{\IEEEauthorrefmark{1}School of Mechatronic Engineering and Automation, Shanghai University, Shanghai, China}
\IEEEauthorblockA{\IEEEauthorrefmark{2}Shanghai Baoshan Shangda General Intelligent Robotics Research Institute, Shanghai, China}
\IEEEauthorblockA{\IEEEauthorrefmark{3}School of Computer Science and Technology, Tongji University, Shanghai, China}
\IEEEauthorblockA{Emails: \{fu23120812, jiafenghuang, yuchenwang25\}@shu.edu.cn}
\IEEEauthorblockA{shengjiezhao@tongji.edu.cn}
\IEEEauthorblockA{Corresponding author: Jiafeng Huang (jiafenghuang@shu.edu.cn)}
}

\begin{document}
\maketitle
\begin{abstract}
Neural radiance fields and 3D Gaussian Splatting assume that each training
image is a sharp and geometrically consistent observation of the scene. Motion blur violates this assumption because a single exposure integrates a continuous range of camera poses. Exposure integration also removes the temporal information needed to recover the corresponding sharp observation. Event
cameras preserve this information at microsecond resolution and therefore
provide a natural complement to conventional images. Existing event-assisted
reconstruction methods predominantly obtain image supervision through analytical
inversion of the Event Double Integral. Learned restoration from frames and
events offers a second prior. Although weaker when used alone, it fails in
different regions and provides complementary evidence. We present JADE-GS,
which formulates the combination of these priors as spatial evidence allocation.
A lightweight Spatial Prior Router predicts a pixelwise allocation using only the
blurry frame and event stream, then fuses the two fixed restorations into
an additional supervision target. The router is trained without
a sharp reference using consistency with the scene under reconstruction and the
measured exposure, and is removed after optimization. Experiments show that
JADE-GS achieves leading perceptual quality on both benchmarks, attains the best
fidelity on the real benchmark, and remains competitive on the synthetic one. It
requires substantially lower training overhead than diffusion-based alternatives
and preserves native 3DGS rendering with no generative decoding at inference.
\end{abstract}

%%%%%%%%%%%%%%%%%%%%%%%%%%%%%%%%%%%%%%%%%%%%%%%%%%%%%%%%%%%%%%%%%%%%%%%%%%%%%%%%%
\section{Introduction}

High-quality novel-view synthesis assumes that each training image is a
geometrically consistent observation of the scene. Both implicit radiance fields
\citep{mildenhall2020,barron2023} and explicit 3D Gaussian Splatting (3DGS)
\citep{kerbl2023} rely on this assumption. Motion blur violates it at capture
time because a single exposure integrates a range of camera poses and suppresses
the high frequency information needed for sharp appearance and consistent
geometry \citep{wang2023,zhao2024,lee2024deblurgs}. The blurred frame alone
cannot reliably recover this intra-exposure structure because the temporal
information has already been integrated away. Event cameras
\citep{lichtsteiner2008,gallego2020} retain it by recording brightness changes
asynchronously at microsecond resolution. They remain informative in low light
and during high speed motion, where a longer RGB exposure only increases blur
\citep{yang2025ahdinet,xu2025mat}. Following established benchmarks, we study
ego-motion blur in a static scene observed by a fast-moving camera.

\begin{figure}[!t]
\centering
\includegraphics[width=0.96\columnwidth]{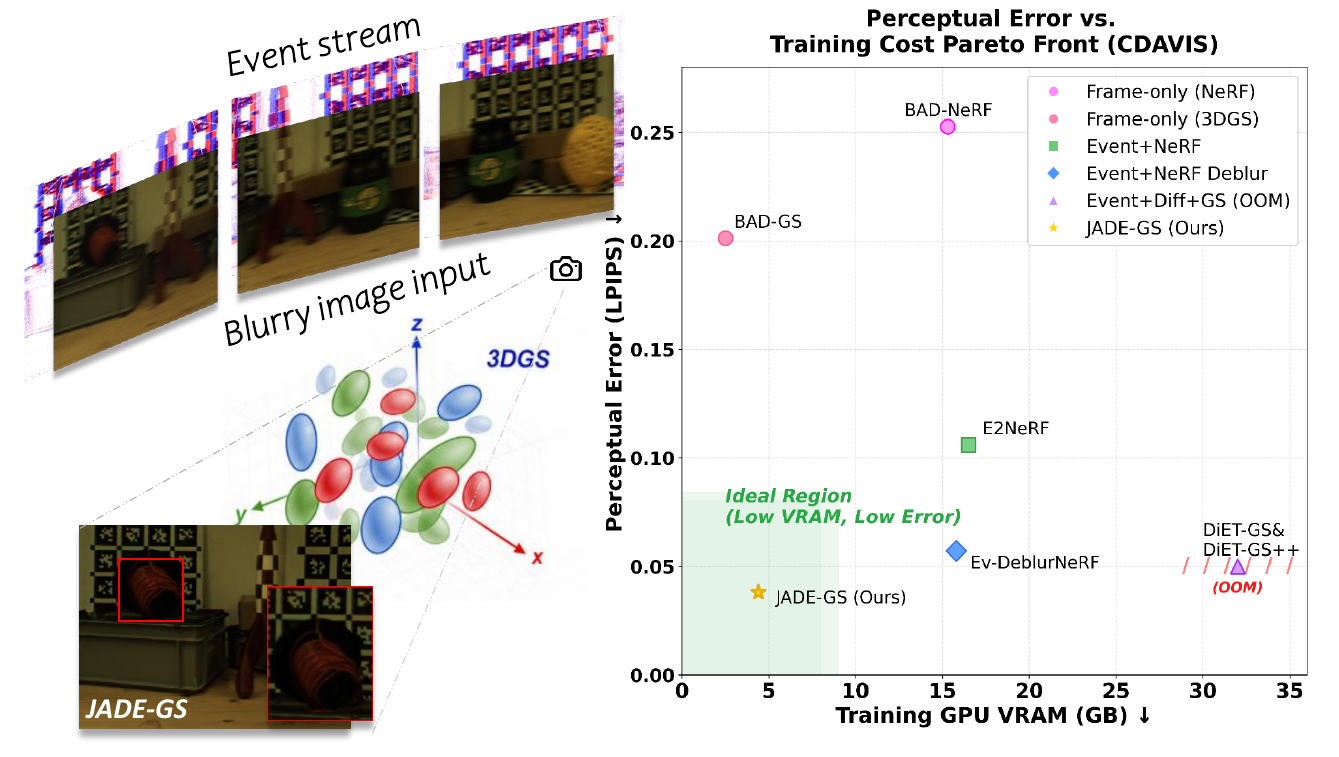}
\caption{The left panel shows reconstruction from blurry frames and event streams
together with a representative JADE-GS result. The right panel shows perceptual
error versus training memory on EvDeblur-CDAVIS. JADE-GS combines strong
perceptual quality with low training memory and native 3DGS rendering. Memory
for baselines with diffusion priors is a lower bound; see the Efficiency
Analysis.}
\label{fig:pareto}
\end{figure}

Most event-assisted reconstruction methods convert the asynchronous event stream
into image supervision through analytical inversion of the Event Double Integral
(EDI) \citep{pan2019edi}. This formulation underlies the EDI loss of
\citet{cannici2024evdeblurnerf}, the EDI-derived constraints of
\citet{lee2025dietgs}, and the mid-exposure EDI target of
\citet{zou2026eventaided}. The analytical prior follows the event generation model
and localizes boundaries supported by events. Its reliability nevertheless
degrades when events are sparse or noisy, allowing integration noise and drift to
propagate into the reconstructed scene. Existing methods address this problem by
either reducing the EDI weight globally during training
\citep{zou2026eventaided} or adding a diffusion prior through score distillation
\citep{lee2025dietgs}. The first strategy suppresses analytical guidance across
the whole image. The second requires generative optimization during training and
a decoder pass for every rendered frame. Both remain centered on the analytical
restoration and do not exploit a separate learned restoration prior.

\begin{figure}[!htbp]
\centering
\includegraphics[width=\columnwidth]{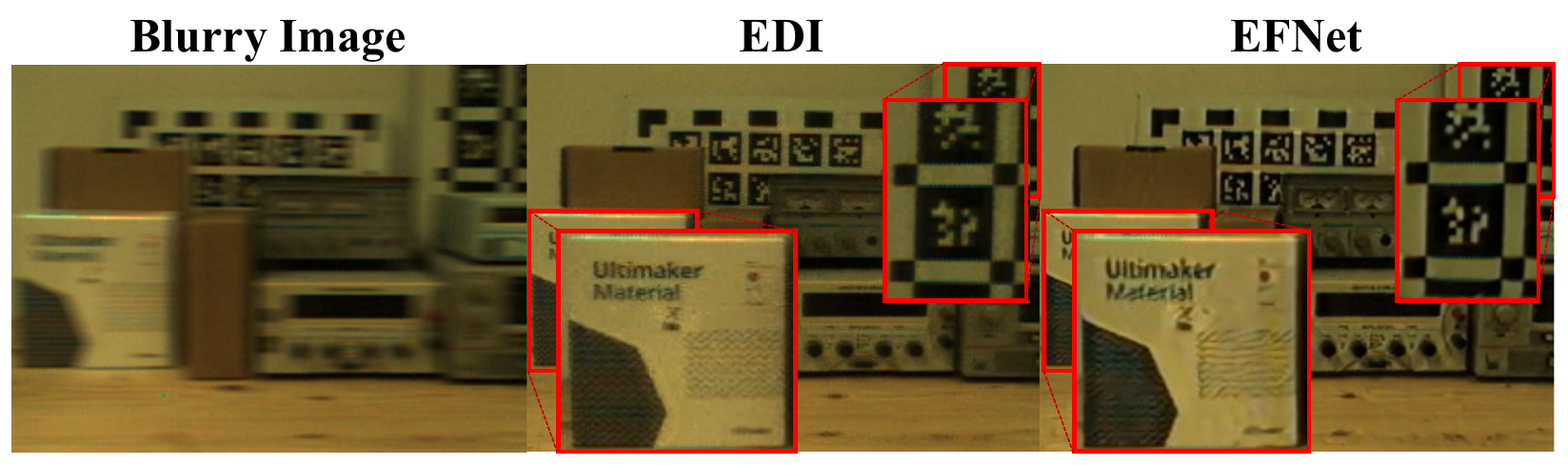}
\caption{Complementary restoration priors. Standalone EDI preserves several
high-contrast boundaries but exhibits integration noise and drift, whereas
standalone EFNet produces cleaner regions but softens some fine edges. JADE-GS
allocates between the two restorations rather than between the observation and
either prior.}
\label{fig:priors}
\end{figure}

Learned restoration from frames and events provides such a second prior. EFNet
\citep{sun2022efnet}, for example, directly restores a sharp image from a blurry
frame and its associated event stream. Although this learned prior is weaker than
analytical inversion when used alone, the two produce complementary errors. EDI
preserves several high contrast boundaries supported by events but may carry
integration noise and drift. EFNet produces cleaner regions but can soften fine
structures, as illustrated in Fig.~\ref{fig:priors}. Uniform combinations do not
make effective use of this complementarity. Fixed weights leave a persistent
conflict between fidelity and perceptual quality, while a learned global weight
converges toward one branch rather than a useful mixture. These effects are
examined in detail in the ablation studies. Since the relative reliability of
the two priors changes across the image, the central question is how their
evidence should be allocated at each pixel.

The allocation should depend on cues that determine local reliability. Event
support indicates where analytical inversion has reliable observational evidence,
while image content carries cues to whether local structure resembles the data
on which the learned restorer was trained. The router therefore reads the
original blurry frame and event stream rather than either restored image.

\begin{figure*}[!htbp]
\centering
\includegraphics[width=0.86\textwidth]{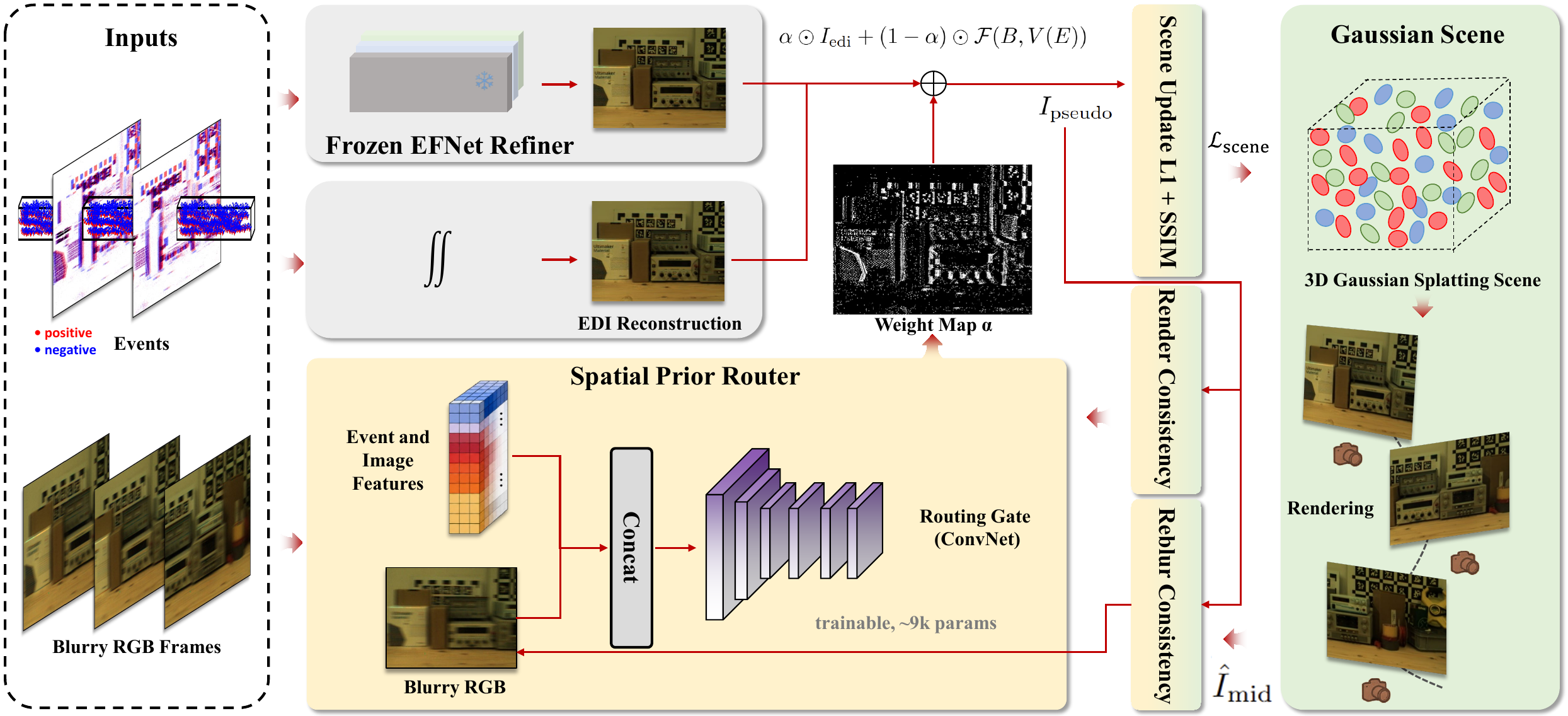}
\caption{Overview of JADE-GS. The analytical EDI branch and the frozen EFNet
refiner $\mathcal{F}$ process the same blurry frame and event stream in parallel.
The Spatial Prior Router $\mathcal{R}_{\theta}$ reads only these measurements and
predicts a pixelwise allocation map that fuses the two restorations into
$I_{\mathrm{pseudo}}$. The fused target complements the inherited analytical
supervision. The router is trained without sharp references through render and
reblur consistency. All routing components are discarded after optimization.}
\label{fig:overview}
\end{figure*}

We present \textbf{JADE-GS}, which treats the combination of analytical and
learned restoration as spatial evidence allocation. As shown in
Fig.~\ref{fig:overview}, the analytical EDI branch and a frozen EFNet refiner run
in parallel. A lightweight Spatial Prior Router predicts a pixelwise allocation
from the original measurements and fuses the two restorations into an additional
supervision target. The inherited analytical supervision of the reconstruction
backbone remains unchanged. The router requires no sharp reference and learns
from consistency signals already available during scene reconstruction. After
optimization, the router and both restoration branches are removed, and novel
views are rendered with the standard 3DGS pipeline.

Our contributions are:
\begin{itemize}
\item We carry the complementarity of analytical event inversion and learned
frame--event restoration into 3D reconstruction, where it has not previously been
exploited. We show that a spatially uniform weight does not exploit it effectively
and formulate their combination as spatial evidence allocation.
\item We introduce a Spatial Prior Router that uses the original measurements to
allocate the two restoration priors at each pixel. It learns without sharp ground
truth from consistency signals already available during reconstruction.
\item Controlled studies isolate the benefit of spatial evidence allocation, and
JADE-GS leads nearly all perceptual metrics on both benchmarks as well as the
fidelity metrics on the real one, at low training overhead and with standard
real-time 3DGS rendering and no generative model at inference.
\end{itemize}

\section{Related Work}

\subsection{Blur-Aware 3D Reconstruction}
Blur-aware radiance-field methods reconstruct sharp novel views by incorporating
finite-exposure image formation into scene optimization. NeRF-based approaches
represent a blurry observation as an integral over latent views or progressively
refine a blur model \citep{ma2022,peng2023pdrf,wang2023}. Related 3DGS methods
optimize explicit Gaussian primitives together with camera motion or view-specific
blur parameters \citep{zhao2024,lee2024deblurgs}. These methods recover blur from
RGB observations alone. Under fast motion, however, an exposure-averaged frame
provides limited temporal evidence for jointly estimating the latent scene and
intra-exposure motion. Event measurements address this ambiguity by directly
recording brightness changes during the exposure.

\subsection{Event-Based Image Deblurring}
Event-based image deblurring follows analytical and learned approaches. EDI
\citep{pan2019edi} reconstructs a latent image from the event-generation model.
It preserves boundaries supported by events, but is sensitive to sparse or noisy
events, contrast-threshold mismatch, and accumulated integration error. Learned
methods such as EFNet \citep{sun2022efnet} and subsequent fusion networks
\citep{yang2024stc,yang2025ahdinet,xu2025mat} combine frame and event features to
model richer spatial and temporal context. They can suppress local artifacts and
complete regions with weak event support, but their behavior depends on the
training data and degradation distribution. These characteristics are individually
documented in 2D restoration, but their complementarity has not been carried into
3D reconstruction. JADE-GS does not introduce another image-restoration
network. Instead, it uses fixed
representatives of the two approaches as candidate supervision for the same 3D
scene.

\subsection{Event-Assisted 3D Reconstruction}
Events have been incorporated into NeRF
\citep{rudnev2023eventnerf,hwang2023evnerf,qi2023e2nerf,cannici2024evdeblurnerf}
and 3DGS
\citep{wu2024evgs,deguchi2024e2gs,weng2024eadeblur,yu2025evagaussians,zahid2025e3dgs}
to support reconstruction under rapid motion and limited exposure. Beyond
optimizing the scene representation, existing methods differ in which additional
components they model or learn. Some derive analytical image or intensity
supervision from events
\citep{cannici2024evdeblurnerf,lee2025dietgs,zou2026eventaided}. Others
additionally estimate the camera trajectory within each exposure
\citep{deng2025ebad,zou2026eventaided,li2025deblursplat,dai2026asyncevgs}, while
EvFlow-GS \citep{an2026evflowgs} replaces analytical latent-image formation with a
learned network. To our knowledge, none makes learnable the allocation of the
analytical prior itself. The learned frame--event
restoration literature offers a second prior, which we are not aware of any
event-guided 3D pipeline using. The likely reason is that on its own the learned
prior is the weaker of the two, as our experiments confirm.

A closely related diffusion-based approach is DiET-GS \citep{lee2025dietgs}, which
retains analytical event supervision and adds a diffusion prior through renoised
score distillation adapted from \citet{lee2024disrnerf}. It reaches high
perceptual quality, but requires generative optimization during training and a
latent decoder pass for every rendered frame. We take the opposite route, making
learnable the per-pixel allocation between two off-the-shelf restorations, and
attenuating the analytical proposal where it fails instead of fading it out
everywhere.

%%%%
\section{Method}

Following the setting adopted by most work in this area
\citep{cannici2024evdeblurnerf,lee2025dietgs,zou2026eventaided}, we supervise the
Gaussian scene with the analytical prior. On top of that, JADE-GS forms a second
target by placing the two restorations side by side and learning only how to weight
them. The analytical branch inverts the event double integral, and the learned
branch is a frozen frame--event refiner; both read the same blurry frame and event
stream, and neither is modified during optimization. A Spatial Prior Router reads those same
measurements and emits a per-pixel weight that fuses the two restorations into a
single target for the Gaussian scene.

Training alternates a routing update with a scene update (Fig.~\ref{fig:overview}).
We fix notation before describing them.

\subsection{Setup and Notation}
\label{sec:setup}
Let $I^{\star}(t)$ denote the latent sharp image at a continuous camera pose
$\pi(t)$, and let $B$ be the observed blurry frame produced by finite-exposure
integration over $[t_s,t_e]$:
\begin{equation}
B = \frac{1}{t_e-t_s}\int_{t_s}^{t_e} I^{\star}(t)\,dt + \epsilon_b,
\label{eq:blur}
\end{equation}
where $\epsilon_b$ absorbs sensor noise, exposure mismatch and residual
misalignment. Following the inherited reconstruction backbone, we approximate this
exposure integral with the same nine sampled camera poses within $[t_s,t_e]$; the
middle sample corresponds to mid-exposure $t_{\mathrm{mid}}=(t_s+t_e)/2$ and defines
the sharp target used throughout the routing and scene updates. The synchronized
event stream is
$E=\{(x_i,y_i,t_i,p_i)\}_{i=1}^{N}$, where each event records a
log-intensity change at pixel $(x_i,y_i)$ that crosses the sensor contrast
threshold, with polarity $p_i\in\{+1,-1\}$.

We denote the Spatial Prior Router by $\mathcal{R}_{\theta}$, the frozen
EFNet refiner by $\mathcal{F}$, and the 3D Gaussian Splatting model by
$\mathcal{G}_{\phi}$. Given $B$ and $E$, the two priors produce their own
restorations and the router fuses them into $I_{\mathrm{pseudo}}$
(Eqs.~\eqref{eq:alpha}--\eqref{eq:fuse}). We write $\theta$ for the routing
parameters and $\phi$ for the Gaussian parameters.

\subsection{Measurement-Conditioned Pixel-Wise Routing}
\label{sec:routing}
The two priors read the same measurements. Let $V(E)=\{V_k\}_{k=1}^{T}$ denote the
normalized signed event voxel built from the event stream with $T$ temporal bins. The
analytical branch computes the mid-exposure EDI reconstruction
$I_{\mathrm{edi}}=\Phi_{\mathrm{edi}}(B,E)$, and the frozen refiner produces
$\mathcal{F}(B,V(E))$, which we treat as an estimate at the same mid-exposure
reference, so both candidates target $t_{\mathrm{mid}}$ before being fused. The
router reads the measurements only,
\begin{equation}
X_{\mathrm{in}} = \big[\,B,\; \Psi(V(E))\,\big],
\label{eq:input}
\end{equation}
where $\Psi$ denotes the polarity-signed temporal projection on which the event
encoder operates. Neither restoration enters $X_{\mathrm{in}}$: this conditions the
allocation on the measurements that determine local event support and image
structure, rather than on artifacts either restoration has already introduced.

The Spatial Prior Router consists of shallow event and image encoders feeding a
three-layer convolutional head, and maps the routing input of Eq.~\eqref{eq:input}
to a per-pixel weight,
\begin{equation}
\alpha=\mathcal{R}_{\theta}(X_{\mathrm{in}}).
\label{eq:alpha}
\end{equation}
A sigmoid on the final layer keeps $\alpha$ within $[0,1]$ at every pixel. The router
emits one scalar value per pixel, and the resulting single-channel map is broadcast
across the three colour channels when the two restorations are fused into the target
that supervises the scene,
\begin{equation}
I_{\mathrm{pseudo}} = \alpha\odot I_{\mathrm{edi}}
                    + (1-\alpha)\odot\mathcal{F}(B,V(E)),
\label{eq:fuse}
\end{equation}
so $\alpha\!\to\!1$ takes the analytical reconstruction and $\alpha\!\to\!0$ takes
the learned one. This target is added to the inherited scene objective rather than
replacing it, so $\alpha$ decides where the added target reasserts the analytical
prior the backbone already supplies and where it instead brings in the learned
restoration.

EFNet's parameters are excluded from optimization, and both restorations are computed
from the measurements alone, so both are constants with respect to $\theta$. For any
routing loss $\mathcal{L}$, the gradient with respect to the allocation at each pixel
is
\begin{equation}
\frac{\partial\mathcal{L}}{\partial\alpha}
=\sum_{c}\frac{\partial\mathcal{L}}{\partial I_{\mathrm{pseudo}}^{(c)}}
\odot\left(I_{\mathrm{edi}}^{(c)}-\mathcal{F}(B,V(E))^{(c)}\right),
\label{eq:allocation_gradient}
\end{equation}
where $c$ indexes the colour channels, and this signal is back-propagated through
Eq.~\eqref{eq:alpha} to update $\theta$. Disagreement between the two restorations
therefore modulates the allocation signal, while changing $\alpha$ has no effect at
pixels where the two restorations coincide, so the router is trained only on the
pixels where the choice matters. Architecture widths are given in the supplement.

\subsection{Alternating Optimization}
\label{sec:routingstep}
\noindent\textbf{Routing step.} No sharp reference is available to supervise $\alpha$, so we build the
supervision out of two signals the optimization already produces. Neither is a
substitute for a sharp image, and neither fails in the same way, which is what
allows them to be combined. Let
$\hat I_{\mathrm{mid}}=\mathcal{G}_{\phi}(\pi(t_{\mathrm{mid}}))$ be the 3DGS
rendering at the middle sampled pose of Eq.~\eqref{eq:blur}. Both terms act on the
fused prediction:
\begin{equation}
\begin{aligned}
\mathcal{L}_{\mathrm{route}}={}&
\underbrace{\bigl\|I_{\mathrm{pseudo}}-\mathrm{sg}(\hat I_{\mathrm{mid}})\bigr\|_1}_{\text{render consistency}}\\
&+\underbrace{\bigl\|\mathcal{B}_{\mathrm{mid}}(I_{\mathrm{pseudo}},V(E))-B\bigr\|_1}_{\text{reblur consistency}}.
\end{aligned}
\label{eq:lroute}
\end{equation}
Here $\mathcal{B}_{\mathrm{mid}}$ is the event-based reblur operator defined in
Eq.~\eqref{eq:reblur} below, and $\mathrm{sg}(\cdot)$ denotes stop-gradient. Both
$\ell_1$ terms are averaged over pixels and colour channels and carry unit weight,
and a single Adam step on $\theta$ follows. The two terms pull from opposite ends
of the pipeline. The rendering comes from a scene representation fitted across
every training view, so it carries evidence that no single exposure holds. The
reblur term runs the same prediction back through the event-based blur model and
requires it to reproduce the frame that was measured. That anchors the map
to the sensor and keeps it from following an inaccurate early scene. Used alone, either term can be
satisfied without sharpening the prediction; together they close the loop from
measurement to scene and back.

\begin{table*}[!htbp]
\centering
\small
\setlength{\tabcolsep}{4.0pt}
\caption{Quantitative comparison on EvDeblur-CDAVIS and EvDeblur-Blender. Full-reference: PSNR, SSIM, VGG-LPIPS. No-reference: MUSIQ, CLIP-IQA. Best in \textbf{bold}, second best \underline{underlined}. Baseline values are quoted from \citet{lee2025dietgs} under the same public protocol. JADE-GS reports a single run for comparability with the single-run baselines; the supplement reports results across seeds.}
\label{tab:main}
\begin{tabular}{ll ccccc ccccc}
\toprule
& & \multicolumn{5}{c}{\textbf{EvDeblur-CDAVIS} (real)} & \multicolumn{5}{c}{\textbf{EvDeblur-Blender} (synthetic)}\\
\cmidrule(lr){3-7}\cmidrule(lr){8-12}
Method & Type & PSNR$\uparrow$ & SSIM$\uparrow$ & LPIPS$\downarrow$ & MUSIQ$\uparrow$ & CLIP-IQA$\uparrow$
             & PSNR$\uparrow$ & SSIM$\uparrow$ & LPIPS$\downarrow$ & MUSIQ$\uparrow$ & CLIP-IQA$\uparrow$\\
\midrule
MPRNet+GS                & 2D+GS & 27.51 & 0.7514 & 0.2013 & 25.12 & 0.2134 & 18.76 & 0.5912 & 0.3545 & 24.12 & 0.2413\\
EDI+GS                   & 2D+GS & 32.95 & 0.8922 & 0.0790 & 40.06 & 0.2008 & 23.69 & 0.7694 & 0.1375 & 55.13 & 0.2751\\
EFNet+GS                 & 2D+GS & 30.97 & 0.8503 & 0.1142 & 38.23 & 0.1934 & 21.03 & 0.6413 & 0.3214 & 35.13 & 0.2314\\
\midrule
BAD-NeRF                 & frame & 28.47 & 0.7981 & 0.2526 & 19.96 & 0.1791 & 19.78 & 0.6381 & 0.2490 & 23.63 & 0.1888\\
BAD-Gaussians            & frame & 29.12 & 0.8129 & 0.2012 & 22.12 & 0.1812 & 22.23 & 0.7213 & 0.2012 & 32.43 & 0.1993\\
\midrule
E$^2$NeRF                & event & 31.54 & 0.8687 & 0.1059 & 38.82 & 0.2235 & 24.54 & 0.7993 & 0.1624 & 47.31 & 0.2129\\
Ev-DeblurNeRF            & event & 32.30 & 0.8827 & 0.0571 & 41.32 & 0.2211 & 24.76 & 0.8038 & 0.1788 & 42.38 & 0.2300\\
DiET-GS                  & event & \underline{34.22} & \underline{0.9223} & \underline{0.0496} & 45.80 & 0.2072 & \textbf{26.69} & \textbf{0.8607} & 0.1064 & 57.67 & 0.2769\\
DiET-GS++                & event & 33.16 & 0.9039 & 0.0502 & \textbf{50.44} & \underline{0.2415} & 26.23 & 0.8478 & \underline{0.1052} & \underline{59.91} & \underline{0.2960}\\
\midrule
\textbf{JADE-GS (ours)} & event & \textbf{34.53} & \textbf{0.9225} & \textbf{0.0382} & \underline{48.03} & \textbf{0.2452} & \underline{26.51} & \underline{0.8602} & \textbf{0.0993} & \textbf{61.37} & \textbf{0.2997}\\
\bottomrule
\end{tabular}
\end{table*}

Let $\mathcal{E}_k=\sum_{j=1}^{k}V_j$ be the cumulative accumulation of the event
voxel and let $m$ denote the last temporal bin accumulated before the exposure
midpoint, so that
$\Delta\mathcal{E}_{k,\mathrm{mid}}=\mathcal{E}_k-\mathcal{E}_m$ refers every latent
intensity to the same middle-exposure reference as $I_{\mathrm{pseudo}}$. The
discrete reblur operator is
\begin{equation}
\mathcal{B}_{\mathrm{mid}}(I,V(E))=I\odot\frac{1}{T}\sum_{k=1}^{T}
\exp\!\left(c\,\Delta\mathcal{E}_{k,\mathrm{mid}}\right),
\label{eq:reblur}
\end{equation}
where the effective contrast scale $c$ absorbs the event-voxel scaling used by the
normalized surrogate. We normalize each exposure window and set $c=1$, so
Eq.~\eqref{eq:reblur} is a scale-free event-consistency surrogate derived from the
EDI model rather than a radiometrically calibrated reblur. Both the routing target
and the surrogate are referenced to mid-exposure, which avoids a temporal mismatch
between them; the supplement gives the resulting per-window contrast scale.

Each term enters Eq.~\eqref{eq:allocation_gradient} through the sign of its own
residual, taken as a subgradient where that residual vanishes, and scaled in the
reblur case by the positive kernel of
Eq.~\eqref{eq:reblur}; we therefore keep both routing objectives pointwise, which
preserves the local attribution of the allocation signal, and reserve SSIM for the
scene update. Because $I_{\mathrm{edi}}$ inverts the same event-based exposure model,
reblur consistency tends to favor the analytical branch where that inversion is
self-consistent, and is not a neutral selector between the two candidates. Render
consistency counterbalances this tendency with evidence the scene representation
accumulates across views, which is why the two terms are used together.

\noindent\textbf{Scene step.} After the routing update, the routing branch is evaluated again with the updated
parameters under no gradient tracking. The resulting $I_{\mathrm{pseudo}}$ is
detached and used as a reconstruction target for the middle-pose rendering in
both RGB and grayscale. With
\begin{equation}
\ell_{\mathrm{rec}}(A,C)=(1-\lambda_{\mathrm{s}})\lVert A-C\rVert_1
+\lambda_{\mathrm{s}}\bigl(1-\mathrm{SSIM}(A,C)\bigr),
\label{eq:rec}
\end{equation}
the term that carries the fused target is
\begin{equation}
\begin{aligned}
\mathcal{L}_{\mathrm{tgt}}={}&
\ell_{\mathrm{rec}}\!\left(\hat I_{\mathrm{mid}},\mathrm{sg}(I_{\mathrm{pseudo}})\right)\\
&+\ell_{\mathrm{rec}}\!\left(g(\hat I_{\mathrm{mid}}),g(\mathrm{sg}(I_{\mathrm{pseudo}}))\right),
\end{aligned}
\label{eq:lstu}
\end{equation}
where $g(\cdot)$ converts RGB to grayscale. All remaining terms of the scene
objective are inherited unchanged from the aligned backbone
\citep{cannici2024evdeblurnerf,lee2025dietgs},
\begin{equation}
\mathcal{L}_{\mathrm{scene}}=\mathcal{L}_{\mathrm{tgt}}+\mathcal{L}_{\mathrm{blur}}
+\mathcal{L}_{\mathrm{edi}}+\mathcal{L}_{\mathrm{sim}}+\mathcal{L}_{\mathrm{int}},
\label{eq:lscene}
\end{equation}
namely a blurry-frame reconstruction term, EDI supervision in grayscale and colour,
an EDI re-simulation term and an intensity term. The supplement lists the weights and
the iteration at which each becomes active.

\noindent\textbf{Schedule.} The nine scene renderings are computed once per
iteration. The routing step then updates $\theta$ only, using a stop-gradient copy of
the mid-exposure rendering, which retains its computation graph for the scene update
that follows, after which the scene step updates $\phi$. Two choices keep the coupling stable: the fused target stays a per-pixel convex
combination of two restorations of the same measurement, and the frozen refiner cannot
drift online while the scene it supervises is still forming. The optimization was
stable across every scene and seed we ran; the supplement reports the trajectories. At inference every routing-side
component is discarded,
leaving the optimized Gaussian scene and the standard 3DGS rendering pipeline.

%%%%%%%%%%%%%%%%%%%%%%%%%%%%%%%%%%%%%%%%%%%%%%%%%%%%%%%%%%%%%%%%%%%%%%%%%%%%%%%%%
\section{Experiments}
\label{sec:exp}

\noindent\textbf{Datasets.} We evaluate on the two public benchmarks introduced by
\citet{cannici2024evdeblurnerf} and used by other works in the field
\citep{lee2025dietgs,zou2026eventaided}. \textbf{EvDeblur-CDAVIS} is real, five
scenes recorded with a Color-DAVIS346 sensor \citep{li2015colordavis}, where the
blur is genuine motion blur rather than a synthetic average of short exposures.
\textbf{EvDeblur-Blender} is synthetic, four Deblur-NeRF scenes \citep{ma2022} whose
events are simulated with ESIM \citep{rebecq2018esim}. We keep the published capture
settings throughout, and the supplement lists them.

\noindent\textbf{Protocol.} All methods use the same public splits and the same
COLMAP-based pose-estimation protocol adopted by prior event-guided baselines.
Following \citet{lee2025dietgs}, JADE-GS keeps these COLMAP poses fixed and
inherits the nine-pose exposure model unchanged. We report PSNR, SSIM and VGG-based
LPIPS \citep{zhang2018} as full-reference metrics, and MUSIQ \citep{ke2021musiq}
and CLIP-IQA \citep{wang2023clipiqa} as no-reference perceptual metrics, following the public DiET-GS++ evaluator with its default configurations.

\noindent\textbf{Baselines.} We compare against offline 2D-deblurring$+$3DGS pipelines (MPRNet+GS, EDI+GS, EFNet+GS), frame-only blur-aware rendering (BAD-NeRF, BAD-Gaussians), and event-guided joint methods (E$^2$NeRF, Ev-DeblurNeRF, DiET-GS, DiET-GS++). Baseline metrics are quoted from \citet{lee2025dietgs} under their identical public protocol. More recent GS-based systems are discussed in the Related Work section but not tabulated: EvaGaussians \citep{yu2025evagaussians}, E-3DGS \citep{zahid2025e3dgs} and EvFlow-GS \citep{an2026evflowgs} evaluate on different protocols, and EBAD-Gaussian \citep{deng2025ebad}, DeblurSplat \citep{li2025deblursplat} and AsyncEvGS \citep{dai2026asyncevgs} have no public implementation at the time of submission.

\begin{table*}[!htbp]
\centering
\small
\setlength{\tabcolsep}{6pt}
\caption{Controlled comparison on CDAVIS, averaged over five scenes under one shared
iteration budget. Rows differ only in how the added target
$\mathcal{L}_{\mathrm{tgt}}$ is formed and all keep the inherited scene objective of
Eq.~\eqref{eq:lscene}, apart from the row marked $\dagger$. Block~(a) holds the
allocation spatially uniform, at a constant or as a single learned scalar; block~(b)
predicts $\alpha$ per pixel and varies which consistency terms train it. $\dagger$:
this row also removes the inherited EDI supervision, so it differs in two ways; its
$0.61$~dB gap to the learned global scalar, whose weight is likewise effectively zero,
isolates what that supervision contributes.}
\label{tab:ablation}
\begin{tabular}{lccc ccc}
\toprule
Configuration & $\alpha$ & Render cons. & Reblur cons. & PSNR$\uparrow$ & SSIM$\uparrow$ & LPIPS$\downarrow$\\
\midrule
\multicolumn{7}{l}{\emph{(a) Spatially uniform allocation}}\\
\quad Equal average, $\tfrac{1}{2}(I_{\mathrm{edi}}{+}\mathcal{F}(B,V(E)))$ & $0.5$ & -- & -- & 33.14 & 0.9098 & 0.0453\\
\quad Analytical restoration only, $I_{\mathrm{edi}}$ & $1$ & -- & -- & 33.64 & 0.9163 & 0.0530\\
\quad Learned global scalar & global & \checkmark & \checkmark & 33.69 & 0.9166 & 0.0404\\
\quad Learned restoration only, $\mathcal{F}(B,V(E))^{\dagger}$ & $0$ & -- & -- & 33.08 & 0.9109 & 0.0445\\
\midrule
\multicolumn{7}{l}{\emph{(b) Learned per-pixel allocation}}\\
\quad Render consistency only     & per-pixel & \checkmark & -- & 33.71 & 0.9167 & 0.0410\\
\quad Reblur consistency only     & per-pixel & -- & \checkmark & 34.05 & 0.9180 & 0.0424\\
\quad \textbf{Full (JADE-GS)}     & per-pixel & \checkmark & \checkmark & \textbf{34.53} & \textbf{0.9225} & \textbf{0.0382}\\
\bottomrule
\end{tabular}
\end{table*}

\noindent\textbf{Implementation details.} All JADE-GS results use a single 25,000-iteration
budget, shared across scenes, metrics and ablation settings. EFNet is initialized from public GoPro-pretrained weights
\citep{sun2022efnet} and kept frozen, and the routing branch introduces $9{,}314$ trainable convolutional
parameters. Our exposure model,
EDI constraints and Gaussian implementation follow Ev-DeblurNeRF and DiET-GS to keep
the comparison pipelines aligned. Loss weights, optimizer settings and the remaining
constants are given in the supplement.

\subsection{Ablation Studies}
\noindent\textbf{Effect of allocation.} Table~\ref{tab:ablation} isolates the
contribution of allocation under identical code and an identical iteration budget, so
the rows differ only in how the added target is formed, apart from the row marked
$\dagger$. Holding the weight spatially uniform leaves the two metrics
pulling against each other. The equal average, $\alpha\!\equiv\!0.5$, gives
$33.14$~dB / $0.0453$; taking the analytical restoration alone, $\alpha\!\equiv\!1$,
raises fidelity to $33.64$~dB but degrades perceptual distance to $0.0530$. Letting
the optimizer choose the weight gives the strongest uniform row, $33.69$~dB /
$0.0404$, and it reaches it by driving $\alpha$ to $1.4\times10^{-5}$ on every
scene. Given one number, the optimizer abandons
the analytical restoration rather than finding a useful global strength.

The learned per-pixel map reaches $34.53$~dB / $0.9225$ / $0.0382$, ahead of every
uniform setting on all three metrics and ahead of the learned global scalar by
$0.84$~dB and $5.4\%$ LPIPS. Both configurations keep the same inherited objective
and the same fusion, and differ only in whether the weight may vary across the
image, so what separates them is where the analytical restoration enters the target
rather than whether it enters at all. The
qualitative counterpart is Fig.~\ref{fig:priors}.

\noindent\textbf{The two consistency terms.} The two terms contribute in the
directions the routing step predicts. Reblur consistency alone gives the higher PSNR,
$34.05$ against $33.71$, because it anchors the prediction to the recorded exposure;
the measured frame carries no high-frequency content of its own, so the same term
suppresses detail and LPIPS is correspondingly worse, $0.0424$ against $0.0410$.
Render consistency behaves in the opposite way, shortening perceptual distance while
the still-inaccurate early scene limits fidelity. Together the two are best on all
three metrics, $34.53$~dB / $0.9225$ / $0.0382$: one supplies a fidelity anchor and
the other high-frequency evidence, and neither substitutes for the other.

\begin{figure*}[!htbp]
\centering
\includegraphics[width=0.78\textwidth]{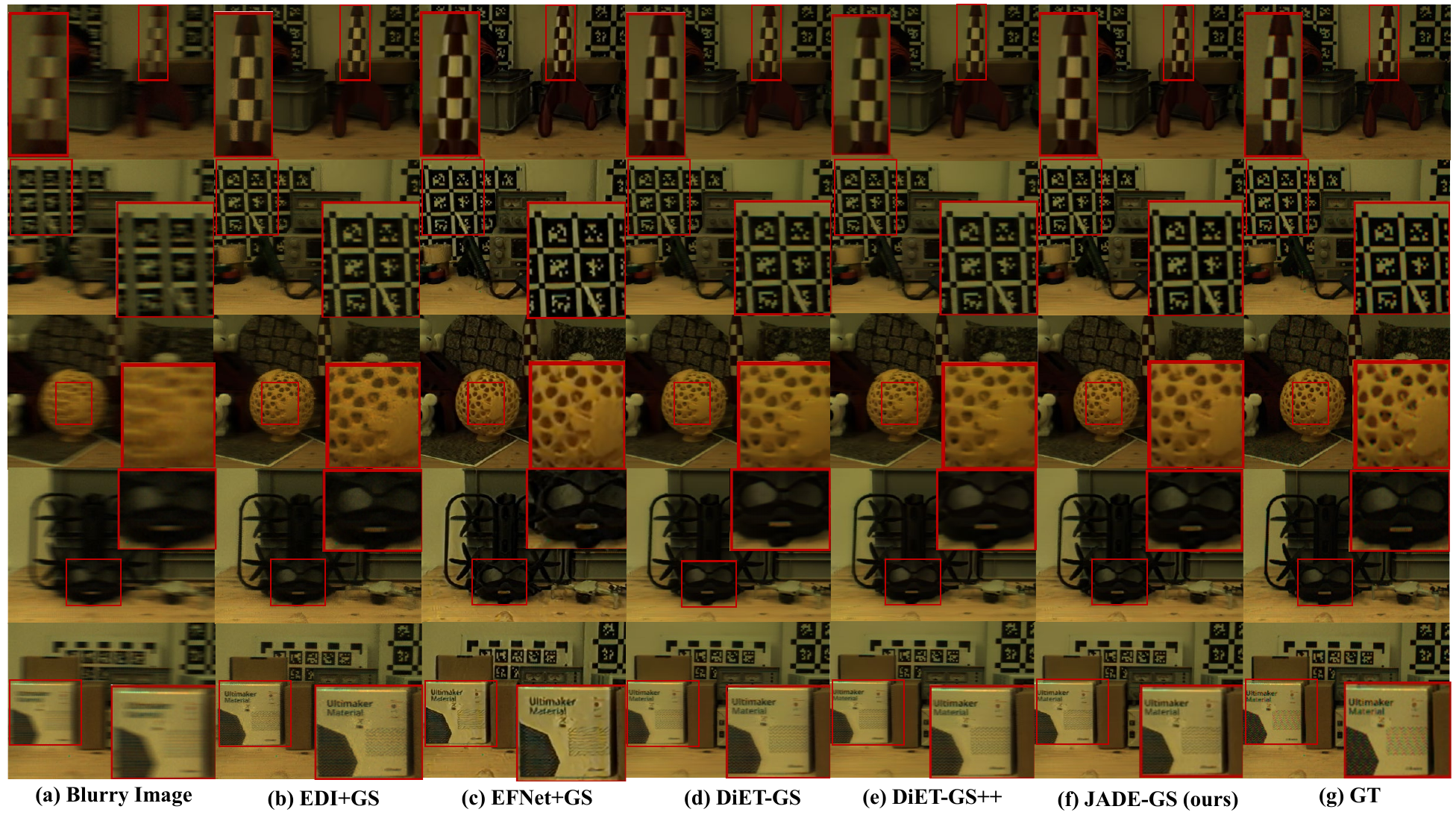}
\caption{Qualitative comparison on the five CDAVIS scenes (one scene per row).
From left to right: blurry input, EDI+GS, EFNet+GS, DiET-GS, DiET-GS++,
JADE-GS (ours), and ground truth. JADE-GS preserves sharper detail on several complex
textures, most visibly in text and repeated patterns; insets enlarge selected
regions.}
\label{fig:qualitative}
\end{figure*}

\begin{figure}[t]
\centering
\includegraphics[width=\columnwidth]{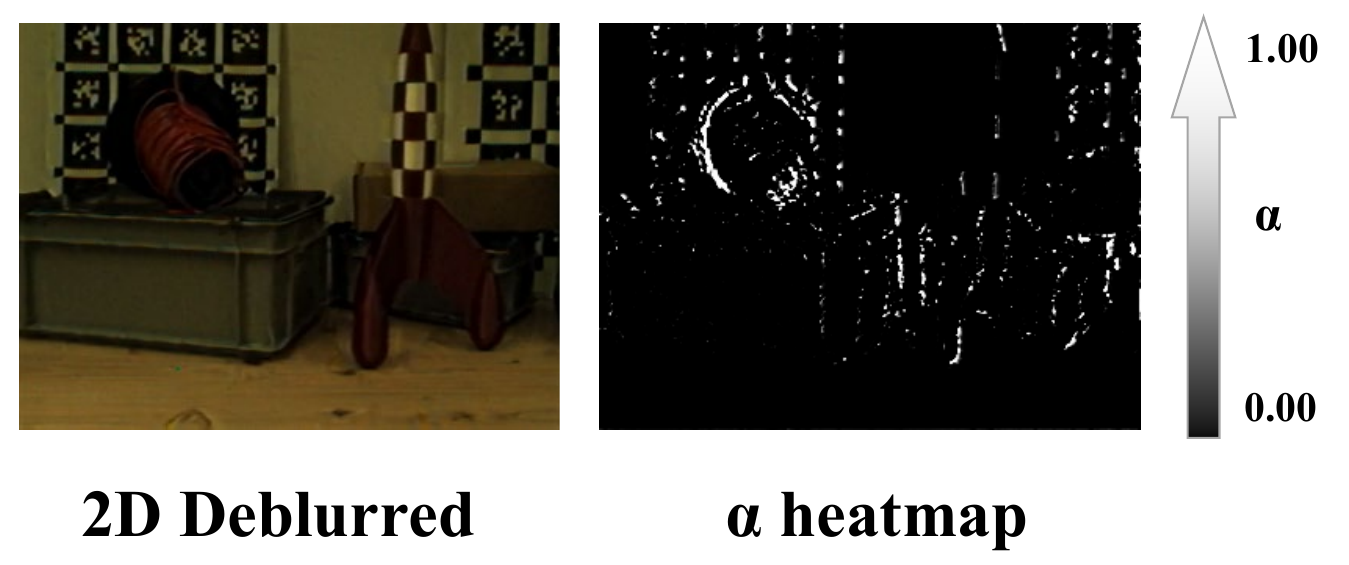}
\caption{Learned routing behavior on a CDAVIS scene. Left: the restored image.
Right: the allocation map $\alpha$, where white denotes $\alpha\approx1$ and takes
the analytical restoration while black denotes $\alpha\approx0$ and takes the
learned one.}
\label{fig:activation}
\end{figure}

\subsection{Comparison with State-of-the-Art Methods}
Table~\ref{tab:main} compares JADE-GS with results published under the same
protocol. Single-frame deblurring followed by 3DGS is
weakest, MPRNet+GS reaching $27.51$~dB on CDAVIS, because nothing in that pipeline
observes the interior of the exposure. Modelling blur formation from RGB alone
represents the exposure integral explicitly but still has only the blurred frame to
invert it, giving $28.47$~dB for BAD-NeRF and $29.12$~dB for BAD-Gaussians. Adding
events lifts the results markedly, to $31.54$~dB for E$^2$NeRF and $32.30$~dB for
Ev-DeblurNeRF. The two offline 2D priors sit at opposite ends of this range,
EDI+GS at $32.95$~dB against $30.97$~dB for EFNet+GS, which indicates why the
analytical prior is the one the field has adopted.

On CDAVIS, JADE-GS cuts LPIPS from $0.0496$ for the strongest baseline to
$0.0382$, a reduction of $23\%$, and this ordering holds on every one of the five
scenes. It also takes the highest PSNR, $34.53$ against $34.22$~dB, again on every
scene, and the highest SSIM, $0.9225$ against $0.9223$. Generated detail need not
align with the reference; routing instead supplies detail that the events support,
placed where the measurement puts it, so perceptual distance can fall without
fidelity being spent.

On the no-reference metrics JADE-GS obtains the best CLIP-IQA and the second-best
MUSIQ. The trade-off behind that ranking is visible in the table: DiET-GS++ gains
$4.64$ MUSIQ over DiET-GS and pays $1.06$~dB of PSNR for it, $33.16$ against
$34.22$. MUSIQ is trained on natural-image quality data and rewards sharp,
high-contrast output, which generative refinement tends to produce whether or not the
texture corresponds to the scene. JADE-GS holds the highest PSNR and SSIM in the
table while taking the best CLIP-IQA and the lowest LPIPS, so its perceptual gain is
not bought with fidelity. On Blender it leads all three perceptual metrics. Events there are
simulated with ESIM, so they are low in noise and follow the event-generation model
closely, which should leave analytical inversion reliable over a larger part of the
image. The gap between the two benchmarks is further evidence that the reliability
of the prior varies with the conditions of capture, which is the premise the method
starts from.

\noindent\textbf{Routing behavior.} Figure~\ref{fig:activation} shows that the
learned map is spatially non-uniform. It is close to zero over most of the frame
and rises on a minority of pixels. This is consistent with the picture the ablation
gives. The analytical prior already constrains the scene everywhere through the
inherited objective, so the added target is most useful where it contributes something
the backbone does not supply. Reasserting the analytical restoration on top of that
pays off only where the learned restoration should not be trusted. Whether a
hand-designed map could play the same role is a question we leave open. No direct
supervision is applied to $\alpha$, and whatever structure it acquires comes entirely
from the two consistency terms.
Figure~\ref{fig:qualitative} shows the corresponding visual effect. JADE-GS
recovers sharper text, edges and repeated textures than the offline 2D+GS
pipelines, and it avoids some of the over-smoothing produced by diffusion
refinement. Reflective packaging remains difficult for every method.

\subsection{Efficiency Analysis}
On the measured RTX~5090 setup, JADE-GS completes training in about
$0.75$~h at a peak of $4.6$~GB per scene (Fig.~\ref{fig:pareto}). \citet{lee2025dietgs} report $9.8$~h for Stage~1 and $10.0$~h for
DiET-GS++ on an RTX~6000. We quote
these runtimes rather than re-measure them, as Stage-1 score distillation exceeds
the $32$~GB available on our GPU. The larger difference is at inference. The routing branch is
discarded, so JADE-GS renders through the unmodified 3DGS rasterizer at $110$~FPS
at the native $346\times260$ resolution. DiET-GS++ and its reduced variant instead
pass every rendered frame through a VAE encoder--decoder, at $1.87$~s per frame.
Measurement protocols are given in the
supplement.

\subsection{Limitations}
JADE-GS targets static scenes under synchronized frame--event capture, the setting of
both benchmarks. Strongly view-dependent reflections remain the hardest case for
every method compared here (Fig.~\ref{fig:qualitative}), and moving objects fall
outside the static-scene assumption. Trajectory refinement \citep{zhao2024},
per-event confidence weighting \citep{fox2024unsupervised} and dynamic
Gaussians \citep{wu2024a,luiten2024} are natural extensions.

%%%%%%%%%%%%%%%%%%%%%%%%%%%%%%%%%%%%%%%%%%%%%%%%%%%%%%%%%%%%%%%%%%%%%%%%%%%%%%%%%
\section{Conclusion}
We introduce JADE-GS, which combines the analytical restoration prior that
event-guided 3D reconstruction relies on with the learned one it has left aside. The
two fail in complementary places, and no spatially uniform weight exploits that, so
JADE-GS allocates between them at each pixel. A Spatial Prior Router sets the weight
from the blurry frame and the event stream alone, supervised only by
scene-rendering consistency and an event-based reblurring constraint, both built from
quantities such a pipeline already computes. On both benchmarks the perceptual advantage previously obtained from a
diffusion prior is reached without one.

%%%%%%%%%%%%%%%%%%%%%%%%%%%%%%%%%%%%%%%%%%%%%%%%%%%%%%%%%%%%%%%%%%%%%%%%%%%%%%%%%

\clearpage
\appendices

\section{Learned Routing Weight $\alpha$}
\label{sec:alpha_analysis}

\begin{figure*}[!t]
\centering
\includegraphics[width=0.98\textwidth]{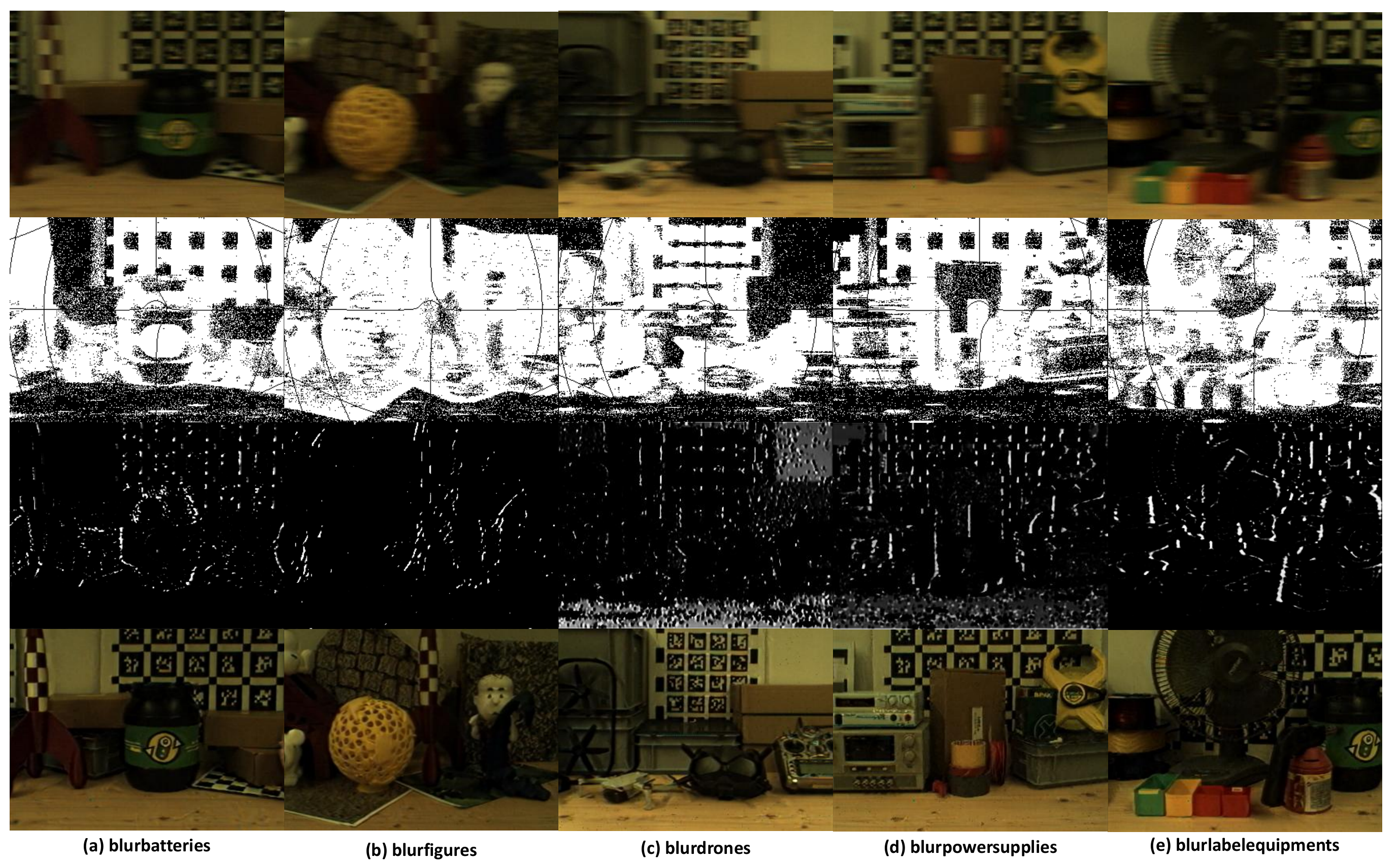}
\caption{Visualization of the learned routing weight $\alpha$ and its relationship to scene content. From top to bottom: the blurry image input, per-pixel event density, $\alpha$ map and 2D deblurred image. Event density is computed as the absolute count of ON/OFF events falling in each pixel during the exposure window. Brighter $\alpha$ means stronger reliance on the analytical EDI prior; darker values mean stronger reliance on the network-restored frame.}
\label{fig:alpha_vis}
\end{figure*}

Figure~\ref{fig:alpha_vis} shows example maps for one representative view from each CDAVIS scene. The routing map is not uniform: it adapts to local blur severity and event density. Table~\ref{tab:alpha_stats} reports scene-level statistics of $\alpha$ and the per-image spatial Pearson correlation between $\alpha$ and event density.

For each training view $i$ we flatten the $\alpha$ map and the count-normalized event density map $c_i$ into vectors $\mathbf{x}=\operatorname{vec}(\alpha_i)$ and $\mathbf{y}=\operatorname{vec}(c_i)$, and compute the spatial correlation
\begin{equation}
r_i = \frac{\sum_j (x_j - \bar{x})(y_j - \bar{y})}{\sqrt{\sum_j (x_j - \bar{x})^2 \sum_j (y_j - \bar{y})^2}} .
\end{equation}
The table reports the mean, standard deviation of $\{r_i\}_{i=1}^{N}$ over all $N$ views in each scene.

\begin{table}[!htbp]
\centering
\caption{Scene-level statistics of the learned routing weight $\alpha$ on CDAVIS at 25k iterations. We report the mean and standard deviation of $\alpha$ over all pixels and views, and the mean spatial Pearson correlation between $\alpha$ and per-pixel event density.}
\label{tab:alpha_stats}
\small
\setlength{\tabcolsep}{6pt}
\begin{tabular}{l|ccc}
\toprule
Scene & $\alpha$ Mean & $\alpha$ Std & Corr.($\alpha$, event density) \\
\midrule
blurbatteries & 0.0268 & 0.0350 & -0.0147 \\
blurfigures & 0.0094 & 0.0237 & 0.0334 \\
blurdrones & 0.0696 & 0.1078 & -0.2269 \\
blurpowersupplies & 0.0598 & 0.0629 & -0.1310 \\
blurlabequipment & 0.0290 & 0.0367 & 0.1518 \\
\bottomrule
\end{tabular}
\end{table}

\paragraph{Discussion.}
At the extremes, $\alpha \approx 1$ means the scene step is driven almost entirely by the analytical EDI reconstruction, while $\alpha \approx 0$ means it follows the refiner output. Visual inspection of Fig.~\ref{fig:alpha_vis} suggests that high-$\alpha$ regions tend to coincide with object boundaries and textured edges rather than uniformly covering the frame. Table~\ref{tab:alpha_stats} confirms that this pattern is not a simple response to raw event counts: Pearson correlations between $\alpha$ and event density are small-magnitude and mixed-sign, ranging from $-0.2269$ to $0.1518$ across the five scenes. The learned weight therefore does not follow a ``more events $\rightarrow$ larger $\alpha$'' rule, nor does it activate on all edges indiscriminately. Instead, it seems to engage the physical EDI prior selectively in local structural regions that are useful for deblurring reconstruction---for instance, boundaries whose geometry is consistent with the motion recorded by events, or whose sharp recovery helps reconstruct the mid-exposure frame. In other words, $\alpha$ encodes not a direct copy of low-level image structure, but an implicit, task-specific judgment of where physical guidance is needed. The mixed-sign and low-magnitude correlations show that the learned allocation is not reducible to a monotonic event-density rule. More importantly, the substantially worse per-view free-map baseline demonstrates that pixel-wise flexibility alone cannot account for the gain; measurement conditioning and cross-view sharing are essential to the proposed formulation.

\section{Per-Scene Quantitative Results}
Table~\ref{tab:supp_per_scene_cdavis} reports the detailed per-scene evaluation on
the five CDAVIS scenes. JADE-GS obtains the best PSNR and the best LPIPS on every
scene, which is the per-scene basis for the claims made in the main paper. Its SSIM
is the best on three scenes and within $0.0022$ of the best value on the remaining
two. On the two no-reference metrics it leads CLIP-IQA on two of five scenes, is
second on two more, and is fourth on the remaining one; on MUSIQ it is second on
all five scenes, with DiET-GS++ ahead where the diffusion prior adds the most
texture.

Baseline entries are reproduced verbatim from Table~7 of
DiET-GS so that they remain comparable with the aggregate figures quoted in the main
paper. We note for transparency that the published CLIP-IQA column for DiET-GS does
not average exactly to the aggregate reported in that work ($0.2095$ per-scene mean
versus $0.2072$ reported); all other columns agree with their aggregates to within
rounding. We do not alter the published values.

\begin{table*}[!t]
\centering
\caption{Per-scene novel-view synthesis on CDAVIS. Metrics: PSNR ($\uparrow$), SSIM ($\uparrow$), LPIPS ($\downarrow$), MUSIQ ($\uparrow$), and CLIP-IQA ($\uparrow$). Best in \textbf{bold}, second best \underline{underlined}. JADE-GS is our method.}
\label{tab:supp_per_scene_cdavis}
\small
\setlength{\tabcolsep}{3.2pt}
\resizebox{\textwidth}{!}{%
\begin{tabular}{l|l|cccccccccc}
\toprule
Scene & Metric & MPRNet+GS & EDI+GS & EFNet+GS & BAD-NeRF & BAD-GS & E$^2$NeRF & Ev-DeblurNeRF & DiET-GS & DiET-GS++ & JADE-GS \\
\midrule
\multirow{5}{*}{{\shortstack{blurbatteries}}}
  & PSNR     $\uparrow$ & 28.42 & 33.11 & 31.30 & 28.29 & 28.73 & 31.49 & 32.63 & \second{34.52} & 33.51 & \best{35.05} \\
  & SSIM     $\uparrow$ & 0.7518 & 0.8994 & 0.8556 & 0.8086 & 0.8217 & 0.8715 & 0.8938 & \second{0.9304} & 0.9118 & \best{0.9309} \\
  & LPIPS    $\downarrow$ & 0.1948 & 0.0613 & 0.0804 & 0.2245 & 0.1651 & 0.0932 & 0.0443 & \second{0.0435} & 0.0444 & \best{0.0346} \\
  & MUSIQ    $\uparrow$ & 22.13 & 37.90 & 35.51 & 17.71 & 20.20 & 37.48 & 42.99 & 45.66 & \best{49.89} & \second{49.48} \\
  & CLIP-IQA $\uparrow$ & 0.2338 & 0.2182 & 0.2293 & 0.1887 & 0.1918 & 0.2445 & 0.2292 & 0.2327 & \second{0.2603} & \best{0.2722} \\
\midrule
\multirow{5}{*}{{\shortstack{blurfigures}}}
  & PSNR     $\uparrow$ & 28.18 & 33.51 & 31.28 & 29.31 & 30.12 & 32.59 & 32.82 & \second{34.89} & 33.86 & \best{35.02} \\
  & SSIM     $\uparrow$ & 0.7311 & 0.8723 & 0.8317 & 0.7703 & 0.7767 & 0.8543 & 0.8577 & \second{0.9049} & 0.8846 & \best{0.9058} \\
  & LPIPS    $\downarrow$ & 0.2146 & 0.0977 & 0.1324 & 0.2935 & 0.2438 & 0.1108 & 0.0687 & \second{0.0600} & 0.0634 & \best{0.0476} \\
  & MUSIQ    $\uparrow$ & 23.45 & 38.48 & 37.13 & 19.50 & 22.13 & 39.47 & 39.70 & 45.37 & \best{51.71} & \second{46.30} \\
  & CLIP-IQA $\uparrow$ & 0.2418 & 0.2384 & 0.2218 & 0.1836 & 0.1898 & 0.2624 & 0.2441 & 0.2584 & \best{0.2955} & \second{0.2888} \\
\midrule
\multirow{5}{*}{{\shortstack{blurdrones}}}
  & PSNR     $\uparrow$ & 27.13 & 33.02 & 31.18 & 28.51 & 29.19 & 31.03 & 31.62 & \second{34.08} & 32.92 & \best{34.30} \\
  & SSIM     $\uparrow$ & 0.7634 & 0.9025 & 0.8617 & 0.8123 & 0.8317 & 0.8780 & 0.8866 & \best{0.9339} & 0.9152 & \second{0.9317} \\
  & LPIPS    $\downarrow$ & 0.2012 & 0.0832 & 0.1293 & 0.2122 & 0.1687 & 0.1075 & 0.0538 & \second{0.0387} & 0.0396 & \best{0.0306} \\
  & MUSIQ    $\uparrow$ & 28.38 & 42.35 & 41.18 & 19.05 & 22.20 & 39.00 & 41.81 & 47.58 & \best{50.17} & \second{49.02} \\
  & CLIP-IQA $\uparrow$ & 0.1718 & 0.1633 & 0.1526 & 0.1723 & 0.1743 & 0.1877 & 0.1773 & 0.1778 & \best{0.2028} & \second{0.2009} \\
\midrule
\multirow{5}{*}{{\shortstack{blurpowersupplies}}}
  & PSNR     $\uparrow$ & 26.37 & 32.10 & 30.92 & 27.35 & 28.38 & 31.06 & 32.05 & \second{33.54} & 32.37 & \best{33.92} \\
  & SSIM     $\uparrow$ & 0.7513 & 0.8955 & 0.8516 & 0.7953 & 0.8071 & 0.8820 & 0.8980 & \second{0.9271} & 0.9108 & \best{0.9293} \\
  & LPIPS    $\downarrow$ & 0.1824 & 0.0657 & 0.1029 & 0.2756 & 0.2247 & 0.0826 & 0.0492 & 0.0460 & \second{0.0459} & \best{0.0310} \\
  & MUSIQ    $\uparrow$ & 31.48 & 46.04 & 44.15 & 24.68 & 24.90 & 45.17 & 47.97 & 50.25 & \best{55.83} & \second{53.17} \\
  & CLIP-IQA $\uparrow$ & 0.2477 & 0.2307 & 0.2219 & 0.1762 & 0.1701 & 0.2373 & \second{0.2501} & 0.2078 & \best{0.2531} & 0.2453 \\
\midrule
\multirow{5}{*}{{\shortstack{blurlabequipment}}}
  & PSNR     $\uparrow$ & 27.47 & 33.00 & 30.18 & 28.89 & 29.19 & 31.51 & 32.36 & \second{34.06} & 33.13 & \best{34.35} \\
  & SSIM     $\uparrow$ & 0.7598 & 0.8911 & 0.8512 & 0.8042 & 0.8276 & 0.8578 & 0.8772 & \best{0.9150} & 0.8971 & \second{0.9148} \\
  & LPIPS    $\downarrow$ & 0.2138 & 0.0871 & 0.1262 & 0.2563 & 0.2037 & 0.1355 & 0.0696 & 0.0599 & \second{0.0575} & \best{0.0471} \\
  & MUSIQ    $\uparrow$ & 20.18 & 35.54 & 33.18 & 18.84 & 21.19 & 32.95 & 34.14 & 40.21 & \best{44.60} & \second{42.18} \\
  & CLIP-IQA $\uparrow$ & 0.1722 & 0.1534 & 0.1418 & 0.1749 & 0.1804 & 0.1854 & \second{0.2048} & 0.1708 & 0.1958 & \best{0.2190} \\
\bottomrule
\end{tabular}
}
\end{table*}

\paragraph{No-reference metric evaluation.}
MUSIQ and CLIP-IQA are computed with IQA-PyTorch following the public DiET-GS++
evaluator, using its default \texttt{musiq} and \texttt{clipiqa} configurations.
Scores are averaged over all test views within each scene and then arithmetically
averaged over the five scenes. Full-reference metrics use the same VGG-based LPIPS variant as
DiET-GS.

\section{Training Budget and Extended Trajectories}
\begin{figure*}[!t]
\centering
\includegraphics[width=0.98\textwidth]{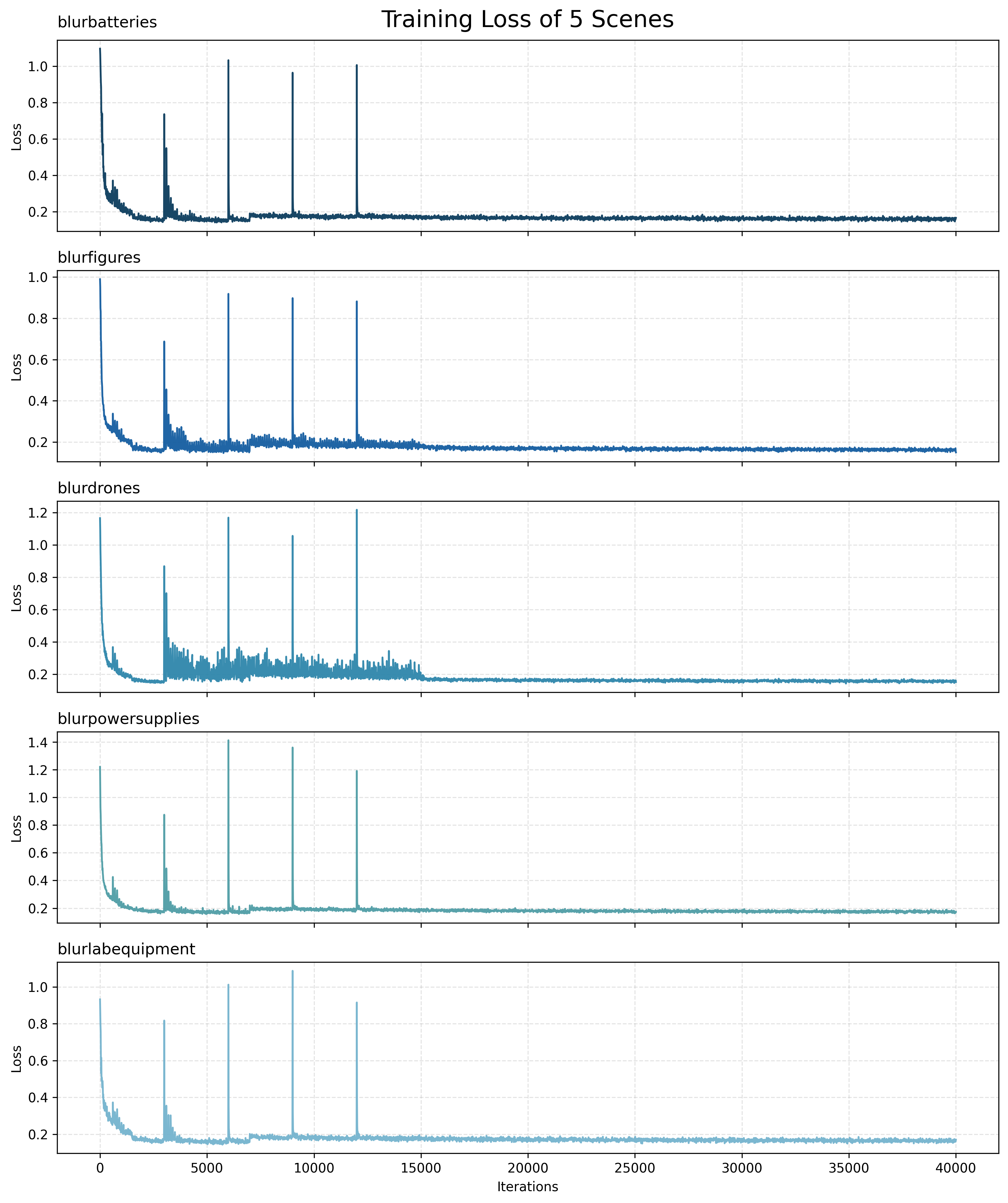}
\caption{Training loss curves for all five CDAVIS scenes. The losses have flattened
by roughly 20k--25k iterations on every scene, which is the criterion used to fix
the reporting budget.}
\label{fig:supp_loss_5scenes}
\end{figure*}

Every JADE-GS result in the main paper uses a fixed budget of 25,000 iterations,
shared across all scenes, metrics and ablation settings. The budget was chosen from
the training objective: as Fig.~\ref{fig:supp_loss_5scenes} shows, the
optimization losses have flattened on all five scenes by this point. Table~\ref{tab:supp_traj_all_scenes} reports
evaluation metrics along trajectories extended to 40k---$1.6\times$ the reporting
budget---purely as a diagnostic of late-stage behavior.

\begin{table*}[!t]
\centering
\caption{Evaluation metrics along training trajectories extended to 40k. The marked 25k column denotes the fixed reporting budget used throughout the main paper; later checkpoints are diagnostic only. Bold denotes the best displayed checkpoint in each row, including ties.}
\label{tab:supp_traj_all_scenes}
\small
\setlength{\tabcolsep}{3.2pt}
\resizebox{\textwidth}{!}{%
\begin{tabular}{l|l|c|c|c|c|c|c|c|c}
\toprule
Scene & Metric & 5k & 10k & 15k & 20k & \textbf{25k$^\dagger$} & 30k & 35k & 40k \\
\midrule
\multirow{3}{*}{blurbatteries}
  & PSNR $\uparrow$ & 34.25 & 34.55 & 34.96 & 34.90 & \best{35.05} & 35.01 & 34.91 & 35.00 \\
  & SSIM $\uparrow$ & 0.9260 & 0.9282 & 0.9293 & 0.9297 & \best{0.9309} & 0.9301 & 0.9280 & 0.9296 \\
  & LPIPS $\downarrow$ & 0.0363 & 0.0361 & 0.0361 & 0.0350 & 0.0346 & 0.0349 & \best{0.0345} & 0.0347 \\
\midrule
\multirow{3}{*}{blurfigures}
  & PSNR $\uparrow$ & 34.36 & 34.93 & 34.81 & 34.97 & \best{35.02} & 34.91 & 34.89 & 34.93 \\
  & SSIM $\uparrow$ & 0.8964 & 0.9029 & 0.9038 & 0.9048 & \best{0.9058} & 0.9049 & 0.9045 & 0.9054 \\
  & LPIPS $\downarrow$ & 0.0512 & 0.0491 & 0.0484 & 0.0479 & 0.0476 & 0.0474 & 0.0468 & \best{0.0466} \\
\midrule
\multirow{3}{*}{blurdrones}
  & PSNR $\uparrow$ & 31.94 & 32.44 & 33.68 & \best{34.34} & 34.30 & 34.23 & 34.13 & 33.85 \\
  & SSIM $\uparrow$ & 0.9212 & 0.9249 & 0.9285 & \best{0.9318} & 0.9317 & 0.9309 & 0.9297 & 0.9265 \\
  & LPIPS $\downarrow$ & 0.0353 & 0.0360 & 0.0311 & 0.0307 & \best{0.0306} & 0.0310 & 0.0315 & 0.0381 \\
\midrule
\multirow{3}{*}{blurpowersupplies}
  & PSNR $\uparrow$ & 32.93 & 33.43 & 33.29 & \best{33.94} & 33.92 & 33.83 & 33.84 & 33.77 \\
  & SSIM $\uparrow$ & 0.9200 & 0.9250 & 0.9239 & 0.9288 & \best{0.9293} & 0.9272 & 0.9272 & 0.9268 \\
  & LPIPS $\downarrow$ & 0.0331 & 0.0320 & 0.0319 & 0.0313 & 0.0310 & 0.0308 & \best{0.0306} & 0.0311 \\
\midrule
\multirow{3}{*}{blurlabequipment}
  & PSNR $\uparrow$ & 33.35 & 33.75 & 34.32 & 34.29 & \best{34.35} & 33.19 & 33.20 & 33.07 \\
  & SSIM $\uparrow$ & 0.9056 & 0.9104 & 0.9137 & 0.9131 & \best{0.9148} & 0.9011 & 0.9014 & 0.8997 \\
  & LPIPS $\downarrow$ & 0.0467 & \best{0.0447} & 0.0463 & 0.0484 & 0.0471 & 0.0668 & 0.0660 & 0.0671 \\
\bottomrule
\end{tabular}
}
\end{table*}

Across the range examined the trajectories are flat or degrade past the 25k reporting budget. From 25k to 40k no scene improves in PSNR; SSIM degrades on every scene, by up to $-0.0151$ on blurlabequipment. LPIPS improves only on blurfigures ($-0.0010$), stays flat on blurbatteries ($+0.0001$) and blurpowersupplies ($+0.0001$), and degrades on blurdrones ($+0.0075$) and blurlabequipment ($+0.0200$). Every scene is therefore already at or past its best operating point by the 25k budget selected from the training objective, which is what makes a single shared budget a reasonable reporting choice rather than a per-scene optimum.

\section{Random-Seed Robustness}
The main text reports a single run so that the state-of-the-art table remains compact; the full per-scene results across three independent random seeds are given in Table~\ref{tab:supp_seed}.

Across all five CDAVIS scenes, the three runs agree to within $0.05$~dB PSNR, $0.0007$~SSIM and $0.0013$~LPIPS at the scene level, and the final averages differ by at most $0.02$~dB PSNR, $0.0003$~SSIM and $0.0004$~LPIPS. The largest single variation is $0.05$~dB PSNR (blurlabequipment), the largest SSIM variation is $0.0007$ (blurbatteries), and the largest LPIPS variation is $0.0013$ (blurfigures); all deviations are confined to the second decimal of PSNR and the third or fourth decimal of the other two metrics. These small deviations indicate that JADE-GS is not sensitive to initialization or training order; the improvements reported in the main table are reproducible rather than the result of a favorable seed.

\begin{table*}[!t]
\centering
\caption{Random-seed robustness of JADE-GS on CDAVIS. We report PSNR
($\uparrow$), SSIM ($\uparrow$), and LPIPS ($\downarrow$) for three independent
runs. S0 is the run currently used in the main paper. Mean and Std are the
across-seed mean and sample standard deviation.}
\label{tab:supp_seed}
\scriptsize
\setlength{\tabcolsep}{2.2pt}
\resizebox{\textwidth}{!}{%
\begin{tabular}{l|ccc|ccc|ccc|ccc|ccc}
\toprule
& \multicolumn{3}{c|}{S0} & \multicolumn{3}{c|}{S1} & \multicolumn{3}{c|}{S2}
& \multicolumn{3}{c|}{Mean} & \multicolumn{3}{c}{Std}\\
Scene & PSNR & SSIM & LPIPS & PSNR & SSIM & LPIPS & PSNR & SSIM & LPIPS
& PSNR & SSIM & LPIPS & PSNR & SSIM & LPIPS\\
\midrule
blurbatteries & 35.05 & 0.9309 & 0.0346 & 35.05 & 0.9302 & 0.0348 & 35.02 & 0.9305 & 0.0349 & 35.04 & 0.9305 & 0.0348 & 0.02 & 0.0004 & 0.0002 \\
blurfigures & 35.02 & 0.9058 & 0.0476 & 35.00 & 0.9058 & 0.0469 & 35.01 & 0.9055 & 0.0463 & 35.01 & 0.9057 & 0.0469 & 0.01 & 0.0002 & 0.0007 \\
blurdrones & 34.30 & 0.9317 & 0.0306 & 34.28 & 0.9317 & 0.0307 & 34.30 & 0.9314 & 0.0307 & 34.29 & 0.9316 & 0.0307 & 0.01 & 0.0002 & 0.0001 \\
blurpowersupplies & 33.92 & 0.9293 & 0.0310 & 33.90 & 0.9292 & 0.0306 & 33.91 & 0.9291 & 0.0304 & 33.91 & 0.9292 & 0.0307 & 0.01 & 0.0001 & 0.0003 \\
blurlabequipment & 34.35 & 0.9148 & 0.0471 & 34.32 & 0.9142 & 0.0463 & 34.37 & 0.9146 & 0.0467 & 34.35 & 0.9145 & 0.0467 & 0.03 & 0.0003 & 0.0004 \\
\textbf{Average} & 34.53 & 0.9225 & 0.0382 & 34.51 & 0.9222 & 0.0379 & 34.52 & 0.9222 & 0.0378 & 34.52 & 0.9223 & 0.0379 & 0.01 & 0.0002 & 0.0002 \\
\bottomrule
\end{tabular}%
}
\end{table*}

\section{Dataset Settings}
We use the two benchmarks of Ev-DeblurNeRF with their published capture settings.
EvDeblur-CDAVIS is real, recorded with a Color-DAVIS346 sensor at $346\times260$.
Each of its five scenes provides 11 to 18 blurry training frames with paired events
under a $1000$~ms exposure, an event threshold of $\Theta=0.25$ for both polarities,
and five ground-truth sharp images from seen and unseen viewpoints. EvDeblur-Blender
is synthetic, derived from four Deblur-NeRF scenes (\emph{factory}, \emph{pool},
\emph{tanabata}, \emph{trolley}) at $600\times400$. Its blurry frames average images
rendered at 1000~FPS over a $40$~ms exposure under a single continuous fast motion,
and its events are simulated with ESIM at $\Theta=0.2$.

\section{Implementation and Optimization Settings}
The training launcher runs all five CDAVIS scenes for 25,000 iterations with grayscale and colour EDI supervision, EDI re-simulation, intensity
supervision, the frozen EFNet branch, and online routing updates enabled.
Table~\ref{tab:supp_settings} lists the principal settings.

\paragraph{Routing branch.}
The routing branch has three parts. A shallow event encoder produces six feature
channels from the polarity-signed temporal projection of the binned voxel. A shallow
image encoder produces eight RGB feature channels. A three-layer convolutional head
is then applied to the resulting 17-channel concatenation, comprising the blurry RGB
frame (3), the event features (6) and the image features (8); neither restoration is
part of this input. The head uses two $3\times3$ convolutions with ReLU followed by a
$1\times1$ sigmoid output, emitting the single-channel routing map $\alpha$, which
then fuses the analytical reconstruction with the output of the frozen refiner. The
two encoders and the head together are the only parameters we add to the DiET-GS
backbone: 3,200 in the event encoder, 2,768 in the image encoder and 3,346 in the
fusion head, for $9{,}314$ newly trainable convolutional parameters in total. The
analytical EDI layer has no parameters. EFNet parameters have
\texttt{requires\_grad=False}. Because both restorations depend only on the
measurements, they are constants with respect to the routing parameters, and the
gradient reaches $\alpha$ directly, scaled pixel-wise by the difference between the
two.

\paragraph{Event representation and the reblur scale.}
For each exposure window the training loop builds a six-bin signed event voxel and
divides it by its maximum absolute value, so the voxel entering both the network and
the reblur operator is scale-free. Eq.~(7) of the main paper is applied to this
normalized voxel with $c=1$. Writing $v_{\max}$ for the per-window normalizer, the
effective contrast scale in raw event-count units is $1/v_{\max}$ and therefore
adapts per window: the operator reproduces the EDI exposure model up to a per-window
scale rather than an absolute radiometric calibration. The per-window normalization
absorbs scale variation in event accumulation, and because the term acts as a
consistency constraint between the two branches rather than as a photometric
measurement, a monotone scale-free surrogate is sufficient and removes one
uncalibrated constant from the objective. With $T=6$ bins the mid-exposure boundary falls after the third
bin, so the post-EFNet restored prediction and the reblur operator share the same
middle-time reference. The routing update is
$\mathcal{L}_{\mathrm{route}}=\mathcal{L}_{\mathrm{3D\rightarrow2D}}+
\mathcal{L}_{\mathrm{reblur}}$ with unit coefficients; no additional
$\lambda_{\mathrm{phy}}$ or $\beta$ multiplier is applied.

\paragraph{What the reblur term rewards.}
The reblur operator is multiplicative, $\mathcal{B}_{\mathrm{mid}}(I,V)=I\odot K(V)$
with $K(V)=\tfrac{1}{T}\sum_k\exp(c\,\Delta\mathcal{E}_{k,\mathrm{mid}})$, so it acts
linearly on the fused target. Writing $r_{\mathrm{edi}}$ and $r_{\mathcal{F}}$ for
the reblur residuals of the two restorations, the residual of the fusion is
$\alpha\,r_{\mathrm{edi}}+(1-\alpha)\,r_{\mathcal{F}}$. The term therefore rewards
whichever prior reblurs closer to the measured frame, and it is not neutral between
them: the analytical reconstruction is itself obtained by inverting the same
exposure model, so wherever that inversion is accurate its residual is small by
construction and the term pushes $\alpha$ up. The bias is not exact, because the
reconstruction uses the calibrated threshold while Eq.~(7) uses the per-window
normalized voxel with $c=1$, but its direction is systematic. We regard this as an
inductive bias of the design rather than a neutral sensor score, and it is one
reason the render-consistency term is needed alongside it: the rendering is
independent of the exposure model and therefore does not share the bias.

\paragraph{Memory.}
The reported peak of 4.6~GB is measured with the refiner evaluated inside the
training loop under gradient checkpointing, a setting inherited from an earlier
variant of the pipeline. Since both restorations depend only on the measurements,
they can equivalently be precomputed once per training view, which removes the
refiner forward pass from the loop entirely.

\paragraph{Inherited schedule constants.}
Two inherited constants (\texttt{warmup} and \texttt{simul\_start}) were tuned by
DiET-GS for a 100k-iteration Stage~1. We keep their absolute values under the shorter
25k budget rather than rescaling them, so the intensity and re-simulation terms
become active proportionally earlier in training.

\begin{table*}[!t]
\centering
\caption{Principal settings in the joint-training configuration.}
\label{tab:supp_settings}
\small
\setlength{\tabcolsep}{6pt}
\begin{tabular}{l|l|c}
\toprule
Component & Code option / quantity & Value \\
\midrule
Main training budget & \texttt{iterations} & 25,000 \\
Event voxel & temporal channels $T$ & 6 \\
Reblur contrast scale & $c$ in Eq.~(7), normalized voxel & 1 \\
Routing optimizer & \texttt{joint\_lr} & $5 \times 10^{-4}$ \\
Scene-step target weight & \texttt{efnet\_weight} & 1 \\
D-SSIM mixture & \texttt{lambda\_dssim} & 0.3 \\
CRF/intensity warm-up & \texttt{warmup} & 1,500 \\
EDI re-simulation start & \texttt{simul\_start} & 7,000 \\
Consistency loss coefficients & consistency / reblur & 1 / 1 \\
Exposure trajectory & inherited sampling & 9 poses; middle reference \\
Routing head & kernels / activations & $3\times3$--$3\times3$--$1\times1$ / ReLU--ReLU--Sigmoid \\
Newly trainable parameters & routing branch only & 9,314 \\
\bottomrule
\end{tabular}
\end{table*}

\section{Ablation: What Each Row Changes}
All rows of the main-paper ablation share the same code, the inherited auxiliary
objectives, the nine-pose exposure model and the same 25k budget. The two priors run
in parallel throughout, and the rows differ only in how the allocation weight
$\alpha$ between them is obtained.

Block~(a) removes the routing branch and holds $\alpha$ spatially uniform. At
$\alpha\!\equiv\!1$ the added target is the analytical reconstruction
$I_{\mathrm{edi}}$; at $\alpha\!\equiv\!0.5$ it is the pixel-wise mean of the two
restorations; and the learned-global-scalar row replaces the map with a single value
the optimizer is free to adjust, which converges to $1.4\times10^{-5}$ on all five
scenes with no variation within a scene. These rows keep the inherited scene
objective of Eq.~(10) unchanged, so they differ from block~(b) only in how $\alpha$
is obtained.

The row marked $\dagger$ is not part of that sweep. It sets $\alpha$ to zero
\emph{and} removes the inherited EDI supervision, so the analytical prior leaves the
pipeline entirely. It therefore differs from the other rows in two ways and is
reported for a different purpose: its gap to the learned global scalar, whose weight
is also effectively zero, isolates what the inherited supervision contributes,
$0.61$~dB of PSNR at essentially the same added target.

Block~(b) enables the Spatial Prior Router and varies which consistency terms train
it. The router reads the blurry frame and the binned event voxel only; neither
restoration enters its input.

A third setting, reported in Sec.~\ref{sec:naive_fusion} below rather than as an
ablation row, applies the two priors in a bare pipeline that feeds the restored
images straight to 3DGS without the inherited auxiliary objectives, which is why its
memory footprint differs (0.5~GB versus 4.6~GB). Its equal average falls below both
single priors there, whereas inside the full pipeline the same average falls between
them; in both settings averaging fails to improve on the better prior.

\paragraph{Parameterization of the routing weight.}
Table~\ref{tab:supp_param} varies how $\alpha$ is parameterized while holding the
rest of the pipeline fixed. Removing the sigmoid so the per-pixel map is
unconstrained lets $\alpha$ leave $[0,1]$ altogether, reaching $-16.6$ at one extreme
and $3.6$ at the other, and doubles LPIPS relative to the full method. The
convex-combination constraint of Eq.~(4) is therefore load-bearing rather than
cosmetic: without it the fused target is no longer a blend of two restorations of the
same measurement, and can leave the intensity range either prior produces. The
no-sigmoid row is single-seed, against three seeds for the full method.

\begin{table*}[!t]
\centering
\setlength{\tabcolsep}{4pt}
\caption{Parameterization of $\alpha$ on CDAVIS, averaged over five scenes at the 25k
budget. The last column reports the form the learned weight takes.}
\label{tab:supp_param}
\small
\begin{tabular}{lcccc}
\toprule
Parameterization of $\alpha$ & PSNR$\uparrow$ & SSIM$\uparrow$ & LPIPS$\downarrow$ & learned $\alpha$\\
\midrule
Per-pixel, no sigmoid        & 32.69 & 0.9033 & 0.0744 & $-0.35$\\
\textbf{Per-pixel, sigmoid (full)} & \textbf{34.53} & \textbf{0.9225} & \textbf{0.0382} & spatially varying\\
\bottomrule
\end{tabular}
\end{table*}

\paragraph{Gradient flow.}
During the scene step the restored image is detached, so the scene-step
reconstruction loss does not update the routing parameters; they are trained only by
render consistency and/or reblur consistency. A row that removed both consistency
terms would leave the routing map without a gradient and would measure its
initialization rather than a learned component; the constant rows of block~(a)
supply the corresponding non-adaptive controls, and the two single-term rows isolate
the two consistency signals.

\section{Training Time, Memory and Rendering Speed}
Table~\ref{tab:supp_efficiency} compares per-scene training cost on
EvDeblur-CDAVIS. Unless noted otherwise, values are measured by us on a single
32~GB RTX~5090 and averaged over five scenes. JADE-GS completes its fixed 25k budget
in 0.75~h at 9.8~it/s and peaks at 4.6~GB.

The DiET-GS runtimes are quoted from the authors' Table~4. It reports 9.8~h for the
single Stage-1 optimization and 10~h for DiET-GS++ (9.8~h Stage~1 plus 0.17~h
Stage~2), measured on one NVIDIA RTX~6000 under a prescribed schedule of 100,000
Stage-1 iterations. The same table lists a reduced variant, DiET-GS++-light, at 1.3~h total; only a
single-scene no-reference score is published for it, so it does not appear on the
quality axis of the main table, and it retains the Stage-2 latent refinement and
therefore the same 1.87~s per-frame rendering cost as DiET-GS++. On a 32~GB card the
Stage-1 score-distillation step exhausts memory, its gradient path running through a
pretrained VAE encoder back to the Gaussian parameters, so the memory entries for
those two rows are lower bounds and we report our own absolute cost rather than a
speedup ratio. The memory difference is attributable to replacing a diffusion refiner with a frozen
lightweight one, not to the routing branch itself. The E$^2$NeRF and Ev-DeblurNeRF
runtimes below are our own measurements and differ from the values reported by
DiET-GS (24.3~h and 3.4~h), which were obtained on different hardware. Offline 2D+GS baselines
are inexpensive because restoration runs once before 3DGS optimization, but they
provide weaker fixed supervision. Throughput is listed for implementation context
only, since an iteration is not comparable across NeRF and 3DGS methods.

\paragraph{Rendering speed.}
At inference the entire routing branch is discarded, so rendering is standard
3DGS rasterization. On a single NVIDIA GeForce RTX~5090, JADE-GS renders at 110~FPS
on average at the native $346\times260$ CDAVIS resolution. The measurement uses a
batch size of one, excludes data loading and image saving, and is averaged over all
test views of the five scenes. For contrast, DiET-GS++ refines every rendered frame through a VAE
encoder--decoder and reports 1.8703~s per frame under its own protocol. Its Stage-1
model, DiET-GS, does render at native 3DGS speed, but it is the weaker of the two on
the no-reference metrics.

\begin{table*}[!t]
\centering
\caption{Training efficiency on EvDeblur-CDAVIS. Our measurements use a single 32~GB RTX~5090; throughput and VRAM are recorded after warm-up. $^\dagger$: runtime quoted from the DiET-GS paper (single NVIDIA RTX~6000). $^\ddagger$: the pipeline exhausts memory at diffusion-prior loading in our 32~GB reproduction, so the listed memory is a lower bound. $^\ast$: fixed 25k budget used for all main-paper results.}
\label{tab:supp_efficiency}
\small
\setlength{\tabcolsep}{1.8pt}
\resizebox{0.65\textwidth}{!}{%
\begin{tabular}{ll|ccc}
\toprule
& & \multicolumn{3}{c}{\textbf{EvDeblur-CDAVIS (real), $346\times260$}} \\
\cmidrule(lr){3-5}
Method & Type & Time (h)$\downarrow$ & Speed (it/s)$\uparrow$ & VRAM (GB)$\downarrow$ \\
\midrule
MPRNet+GS            & 2D+GS & $\sim$0.5\,h & 75$\sim$80 & 0.3 \\
EDI+GS               & 2D+GS & $\sim$0.5\,h & 75$\sim$80 & 0.3 \\
EFNet+GS             & 2D+GS & $\sim$0.5\,h & 75$\sim$80 & 0.3 \\
\midrule
BAD-NeRF  & frame & $\sim$10\,h & 5.66 & 15.3 \\
BAD-Gaussians  & frame & $\sim$0.5\,h & 27.6 & 2.1 \\
\midrule
E$^2$NeRF     & event & $\sim$10\,h  & 5.4 &16.5 \\
Ev-DeblurNeRF  & event & $\sim$6\,h  & 2.2 & 15.8 \\
DiET-GS$^{\dagger,\ddagger}$ & event & 9.8\,h & -- & $>32$ \\
DiET-GS++$^{\dagger,\ddagger}$ & event & 10\,h & -- & $>32$ \\
\textbf{JADE-GS (ours)} & event & 0.75\,h$^{\ast}$ & 9.8 & 4.6 \\
\bottomrule
\end{tabular}%
}
\end{table*}

\section{Offline Combination of the Two Priors}
\label{sec:naive_fusion}

This experiment is reported here in full and is not tabulated in the main paper.
Each configuration precomputes a restored
image and uses it as fixed supervision for 3DGS. The two single-prior rows use
$I_{\mathrm{edi}}$ and $\mathcal{F}(B,V(E))$ respectively, and the third takes their
equal average, $\tfrac{1}{2}(I_{\mathrm{edi}}+\mathcal{F}(B,V(E)))$. No routing is
involved, and no inherited auxiliary objective is active.

The equal average is the weakest configuration in PSNR and SSIM on both benchmarks,
losing $6.39$~dB relative to EDI alone and $2.59$~dB relative to EFNet alone on
CDAVIS, and $2.67$~dB relative to EDI on Blender. Averaging two restorations whose
errors are not aligned spreads each prior's failures across the whole image instead
of confining them to the regions where that prior is weak. The same ordering appears
inside the full pipeline, where the equal average of block~(a) again fails to improve
on the better of the two priors.

\begin{table}[!htbp]
\centering
\caption{Offline application of the two restoration priors. All values are measured
by us under the same optimization protocol and evaluation pipeline used for this analysis and are therefore
not directly comparable to the baseline rows in Table~1 of the main paper, which are quoted from DiET-GS.}
\label{tab:naive_fusion}
\small
\setlength{\tabcolsep}{5pt}
\begin{tabular}{l|ccc}
\toprule
Method & PSNR$\uparrow$ & SSIM$\uparrow$ & LPIPS$\downarrow$ \\
\midrule
\multicolumn{4}{c}{EvDeblur-CDAVIS} \\
EDI$\rightarrow$GS & 33.35 & 0.9006 & 0.0685 \\
EFNet$\rightarrow$GS & 29.55 & 0.8493 & 0.0711 \\
$\tfrac{1}{2}($EDI$+$EFNet$)\rightarrow$GS & 26.96 & 0.8067 & 0.1055 \\
\midrule
\multicolumn{4}{c}{EvDeblur-Blender} \\
EDI$\rightarrow$GS & 25.31 & 0.8301 & 0.1092 \\
EFNet$\rightarrow$GS & 24.25 & 0.8018 & 0.1673 \\
$\tfrac{1}{2}($EDI$+$EFNet$)\rightarrow$GS & 22.64 & 0.7926 & 0.1242 \\
\bottomrule
\end{tabular}
\end{table}

\section{Per-View Pixel Router Baseline}
\label{sec:per_view_pixel_router}

To verify that the gains of JADE-GS come from the learned multimodal router rather than from the mere presence of a per-pixel blending weight during training, we test a \textbf{Per-View Pixel Router} baseline. It keeps the same two priors and the same render-plus-reblur losses, but replaces the shared Spatial Prior Router with one independent $\alpha \in [0,1]^{H \times W}$ per training view. This alpha map is a free \texttt{nn.Parameter}: it receives no input, is not shared across views, and is trained solely through the render and reblur consistency losses. The baseline therefore has at least as much per-view freedom as the full router, yet it removes both measurement conditioning and cross-view parameter sharing.

\begin{table*}[!t]
\centering
\caption{Per-View Pixel Router baseline on EvDeblur-CDAVIS. The baseline replaces the shared measurement-conditioned Spatial Prior Router with one independent learnable alpha map per training view.}
\label{tab:per_view_pixel_router}
\small
\setlength{\tabcolsep}{6pt}
\begin{tabular}{l|ccc}
\toprule
Method & PSNR$\uparrow$ & SSIM$\uparrow$ & LPIPS$\downarrow$ \\
\midrule
Per-View Pixel Router (per-view $\alpha$, no input) & 32.16 & 0.8956 & 0.0967 \\
\textbf{JADE-GS (shared measurement-conditioned router)} & \textbf{34.53} & \textbf{0.9225} & \textbf{0.0382} \\
\bottomrule
\end{tabular}
\end{table*}

The baseline drops by $2.37$~dB PSNR and falls well behind in SSIM and LPIPS (Table~\ref{tab:per_view_pixel_router}). This gap shows that the gain is not simply a consequence of adding a learnable per-pixel blending weight: the Per-View Pixel Router actually introduces far more trainable parameters than the shared Spatial Prior Router (one free alpha map per training view versus $9{,}314$ shared convolutional parameters), yet it performs substantially worse. What matters is that the Spatial Prior Router reads the measurements (raw blur and event voxel) and shares its parameters across views, so the allocation policy generalizes rather than overfitting each exposure. Without both measurement conditioning and cross-view sharing, the larger per-view representation merely memorizes each training view, confirming that the Spatial Prior Router plays a specific and necessary role.

\bibliographystyle{IEEEtran}
\bibliography{jade-gs}

\end{document}